\documentclass[10pt,twocolumn,letterpaper]{article}

\usepackage[pagenumbers]{cvpr} 

\usepackage{graphicx}
\usepackage{amsmath}
\usepackage{amssymb}
\usepackage{booktabs}
\usepackage{multicol}
\usepackage{multirow}
\usepackage{float}
\usepackage{color}

\usepackage[pagebackref,breaklinks,colorlinks]{hyperref}

\usepackage[capitalize]{cleveref}
\crefname{section}{Sec.}{Secs.}
\Crefname{section}{Section}{Sections}
\Crefname{table}{Table}{Tables}
\crefname{table}{Tab.}{Tabs.}

\def\cvprPaperID{8799} 
\def\confName{CVPR}
\def\confYear{2023}
\usepackage{booktabs}

\begin{document}

\title{ChatCAD: Interactive Computer-Aided Diagnosis on Medical Image using Large Language Models}

\author{Sheng Wang$^{1,2,3}$\qquad 
Zihao Zhao$^{1}$ \qquad Xi Ouyang$^{3}$ \qquad Qian Wang$^{1}$ \qquad Dinggang Shen$^{1,3}$\\
$^1$ShanghaiTech University \qquad $^2$Shanghai Jiao Tong University \qquad $^3$United Imaging Intelligence
}
\twocolumn[{%
\renewcommand\twocolumn[1][]{#1}%
\maketitle
\begin{center}
    \centering
    \captionsetup{type=figure}
    \includegraphics[width=1\textwidth]{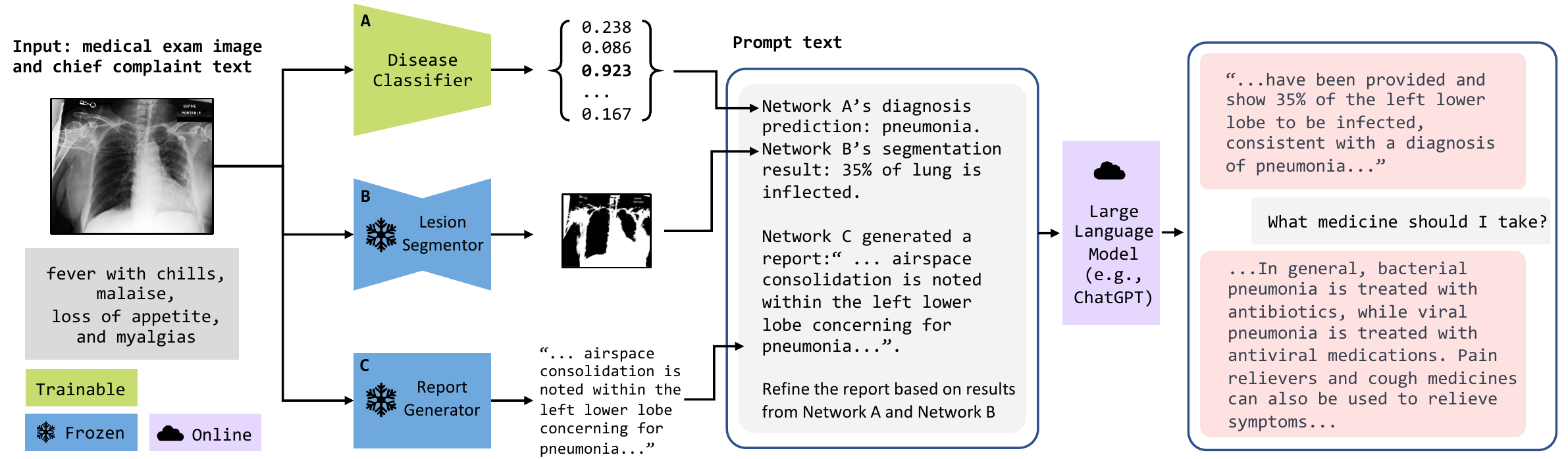}
    \captionof{figure}{Overview of our proposed strategy. 
    The image is processed by various networks to generate diverse outputs, which are then transformed into text descriptions. The descriptions, served as a link between visual and linguistic information, are combined as inputs to a large language model (LLM). With its ability to reason and its knowledge of medical field, the LLM can provide a condensed report and offer interactive explanations and medical recommendations based on the given image.}
    \label{overview}
\end{center}%
}]

\begin{abstract}
Large language models (LLMs) have recently demonstrated their potential in clinical applications, providing valuable medical knowledge and advice. For example, a large dialog LLM like ChatGPT has successfully passed part of the US medical licensing exam.
However, LLMs currently have difficulty processing images, making it challenging to interpret information from medical images, which are rich in information that supports clinical decisions.
On the other hand, computer-aided diagnosis (CAD) networks for medical images have seen significant success in the medical field by using advanced deep-learning algorithms to support clinical decision-making.
This paper presents a method for integrating LLMs into medical-image CAD networks.
The proposed framework uses LLMs to enhance the output of multiple CAD networks, such as diagnosis networks, lesion segmentation networks, and report generation networks, by summarizing and reorganizing the information presented in natural language text format. 
The goal is to merge the strengths of LLMs' medical domain knowledge and logical reasoning with the vision understanding capability of existing medical-image CAD models to create a more user-friendly and understandable system for patients compared to conventional CAD systems. In the future, LLM's medical knowledge can be also used to improve the performance of vision-based medical-image CAD models.
\end{abstract}

\section{Introduction}
Large Language Models (LLMs) are advanced artificial intelligence systems that have been trained on vast amounts of text data~\cite{devlin2018bert,radford2018improving}. These models use deep learning techniques to generate human-like responses, making them useful for a variety of tasks such as language translation, question answering, and text generation.
\begin{figure*}[!ht]
    \centering
    \includegraphics[width=1\textwidth]{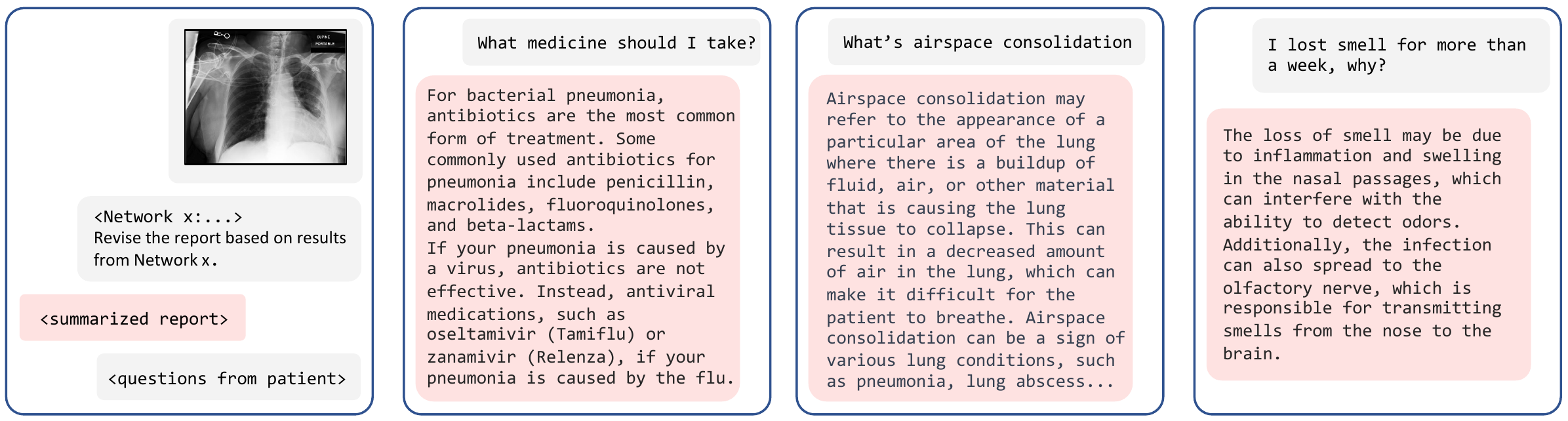}
    \caption{Interactive CAD with LLMs. This example uses the ChatGPT as LLM.}
    \label{usecases}
\end{figure*}
LLMs like OpenAI's GPT-3~\cite{brown2020language} have shown remarkable results in natural language processing and have the potential to revolutionize various industries, including marketing, education, and customer service.
The ability of LLMs to process and understand large amounts of data has made them highly sought after for solving complex problems.
In the medical domain, LLMs have demonstrated their potential as valuable tools for providing medical knowledge and advice. For instance, a large dialog-based LLM, such as ChatGPT~\cite{OpenAI2023ChatGPT}, has demonstrated remarkable results in a critical evaluation of its medical knowledge. ChatGPT has successfully passed part of the US medical licensing exams, showcasing its potential to augment medical professionals in delivering care.
Inspired by their remarkable progress in natural language processing, it is an interesting topic to integrate the LLMs to understand visual information in computer vision tasks. 
Processing images involves understanding the spatial relationships between objects, recognizing patterns and textures, and extracting features that describe the objects in an image. These tasks require a deep understanding of visual information, which is challenging for LLMs that have been primarily trained on text data. 
This limitation presents a major challenge in the medical field, where images play a crucial role in supporting clinical decisions. Medical images, such as X-rays, CT scans and MRIs, are rich in information that can provide critical insights into a patient's condition. However, LLMs currently struggle to interpret and extract information from these images, limiting their ability to fully support clinical decision-making processes.

As the ``pure" computer vision method, medical-image computer-aided diagnosis (CAD) networks have achieved significant success in supporting clinical decision-making processes in the medical field~\cite{shen2017deep}. These networks leverage advanced deep learning algorithms to analyze medical images and provide valuable insights to support clinical decision-making.
CAD networks have been designed specifically to handle the complexities of visual information in medical images, making them well-suited for tasks such as disease diagnosis~\cite{wang2022follow}, lesion segmentation~\cite{zhao2023rcps}, and report generation. These networks have been trained on large amounts of medical image data, allowing them to learn to recognize complex patterns and relationships in visual information that are specific to the medical field.

The aim of this paper is to provide a scheme that combines the strength of LLMs and CAD models.
In this scheme, namely ChatCAD, the image is first fed into multiple networks, i.e., an image classification network, a lesion segmentation network, and a report generation network as depicted in Figure~\ref{overview}. 
The results produced by classification or segmentation are a vector or a mask, which can not be understood by LLMs. 
Therefore, we transform these results into the text representation form as shown in the middle panel of Figure~\ref{overview}. These text-form results will then be concatenated together as a prompt ``\textit{Revise the report based on results from Network A and Network B}'' for the LLM.
The LLM then summarizes the results from all the CAD networks. 
As the example in this figure, the refined report combines the findings from all three networks to provide a clear and concise summary of the patient's condition, highlighting the presence of pneumonia and the extent of the infection in the left lower lobe. 
In this way, the LLM could correct errors in the generated report based on the results from CAD networks. 
Our experiment shows that our scheme could improve the diagnosis performance score of the state-of-the-art report generation methods by $16.42\%$. A major benefit of our approach is the utilization of LLM's robust logical reasoning capabilities to combine various decisions from multiple models. This allows us to fine-tune each model individually. For instance, in response to an emergency outbreak such as COVID-19, we can add a pneumonia classification model (differentiating between community-acquired pneumonia and COVID-19~\cite{ouyang2020dual}) using very few cases without affecting the other models. Since classifiers are usually less data-hungry than other models, we mark it with ``trainable" (green) in Figure~\ref{overview}.

Another advantage of bootstraping LLMs to CAD models is that their extensive and robust medical knowledge can be leveraged to provide interactive explanations and medical advice as we illustrate on Figure~\ref{usecases}. For example, based on an image and generated report, patients can inquire about appropriate treatment options (second panel) or define medical terms such as ``airspace consolidation" (third panel). 
Or with patient's chief complaint (forth panel), LLMs can explain why such symptom happens.
In this manner, patients can gain a deeper understanding of their symptoms, diagnosis, and treatment more efficiently. It can efficiently help patients to reduce consultation costs with clinical experts.
As the performances of CAD models and LLMs become increasingly improved in the future, the proposed scheme has the potential to improve the quality of radiology reports and enhance the feasibility of online healthcare services.

\section{Related Works}
\begin{figure*}[!ht]
    \centering
    \includegraphics[width=1\textwidth]{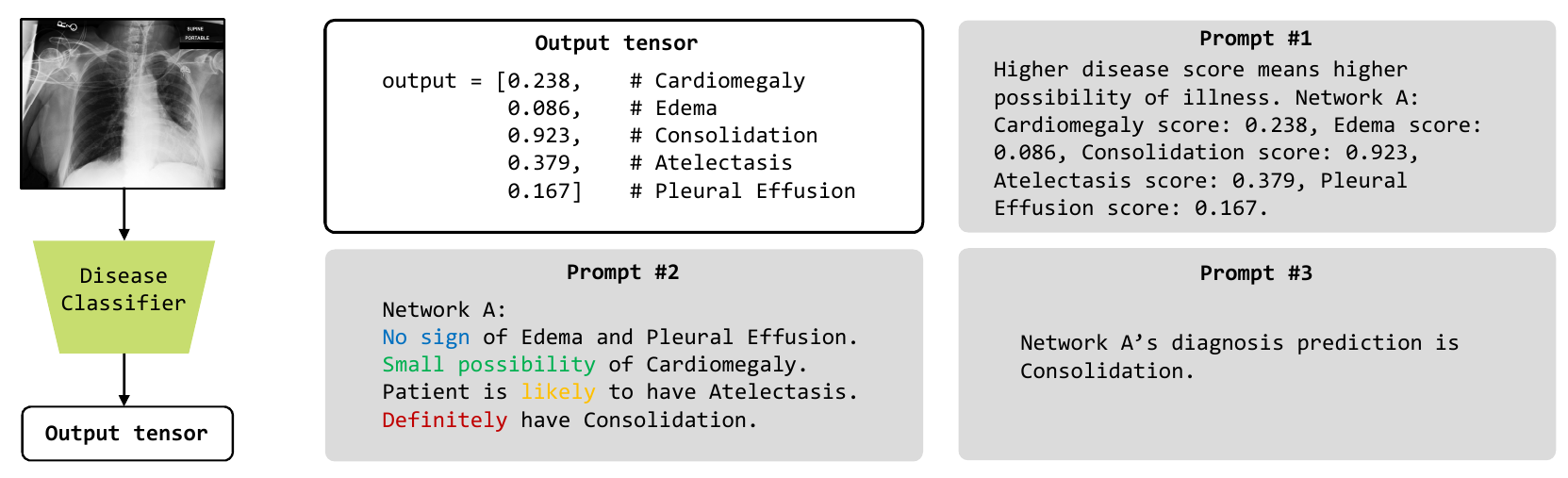}
    \caption{Prompts that bridge between tensor and text. We show three different prompt designs.}
    \label{prompt}
\end{figure*}
\subsection{Large Language Models}
Recent advances in Transformer architecture~\cite{vaswani2017attention} and computing power have enabled the training of large language models with billions of parameters, leading to a significant improvement in their ability to summarize, translate, predict and generate human-like text~\cite{brown2020language,singhal2022large,raffel2020exploring}. 

Several domain-specific LLMs have been developed using general-purpose model weight and training schemes. BioBERT~\cite{lee2020biobert} and PubMedBERT~\cite{gu2021domain} are examples of BERT~\cite{devlin2018bert} models trained on PubMed for biomedical data, while ClinicalBERT~\cite{alsentzer2019publicly} was further trained on the MIMIC dataset and outperformed its predecessor. Med-PaLM~\cite{singhal2022large} was developed in late 2022 using curated biomedical corpora and human feedback, and showed promising results, including a 67.6\% accuracy on the MedQA exam. ChatGPT, which was not given supplementary medical training, passed all three parts of the USMLE and achieved over 50\% accuracy across all exams and surpassed 60\% accuracy in the majority of them~\cite{kung2022performance}.

\subsection{Vision-Language Model}
A popular method of converting visual information into language is through image captioning. Deep learning-based image caption models~\cite{you2016image,herdade2019image} can generate descriptive and coherent captions using large datasets such as Microsoft COCO and Flickr 30K. In medical image analysis, image captioning methods are employed to generate exam image reports.
For example, Li et al.\cite{liu2019clinically} implement explicit medical abnormality graph learning for report generation. Zhang et al.\cite{zhang2020radiologyConGraph} utilize a pre-constructed knowledge graph based on disease topics, respectively. Another line of research~\cite{chen-acl-2021-r2gencmn,wang2022crossPro} learns cross-modal patterns using self-attention architecture. The recent emergence of foundation models with more clinical knowledge holds promise as a potential future direction.


Recently, with the increase in model size, advances in the field have shifted towards Vision-Language Pretraining (VLP) and utilizing pre-trained models. CLIP~\cite{radford2021learning} merges visual and language information into a shared feature space, setting new state-of-the-art performance on various downstream tasks. Frozen~\cite{tsimpoukelli2021multimodal} fine-tunes an image encoder, whose outputs serve as soft prompts for the language model. Flamingo~\cite{alayrac2022flamingo} introduces cross-attention layers into the LLM to incorporate visual features, pre-training these new layers on billions of image-text pairs.

\section{Method}

\subsection{Bridge between Image and Text}
\begin{figure}[!ht]
    \centering
    \includegraphics[width=0.47\textwidth]{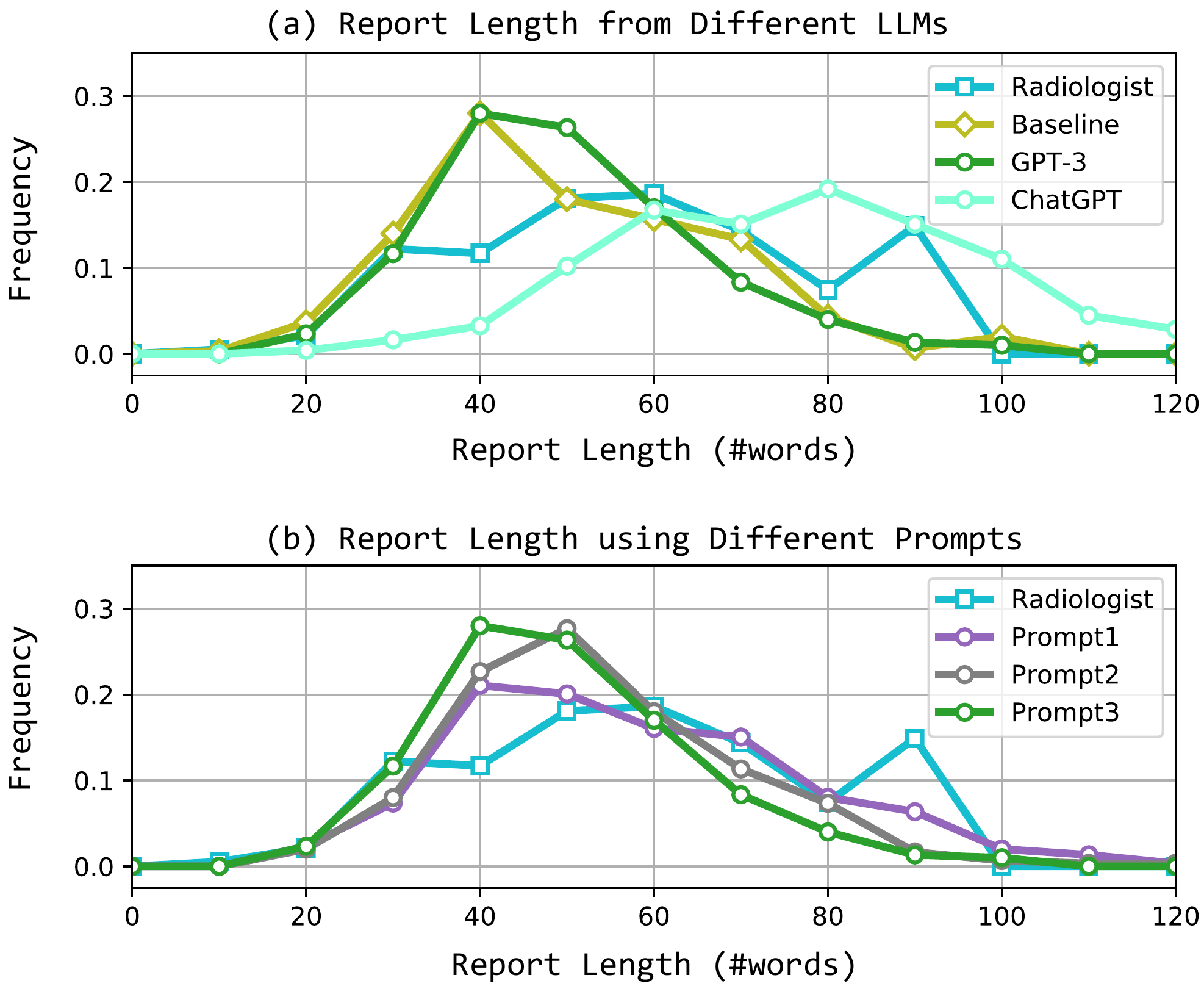}
    \caption{Length comparison of generated reports.}
    \label{length}
\end{figure}
The key idea is to utilize the powerful logical reasoning capabilities of the LLMs to make more robust disease diagnosis for medical images.
Therefore, we need to build a bridge to translate medical images into texts as inputs for the LLM.
Our strategy is straightforward: 1) Feed exam images (e.g., X-Ray) into trained CAD models to obtain outputs; 2) Translate these outputs (typically tensors) into natural language; 3) Use language models to summarize the results and make a final conclusion; 4) Based on the results from visual models and pre-trained medical knowledge in the language models, engage in conversation about symptoms, diagnosis, and treatment. In this section, we mainly discuss the details of our proposed scheme.

\begin{figure*}[!ht]
    \centering
    \includegraphics[width=1\textwidth]{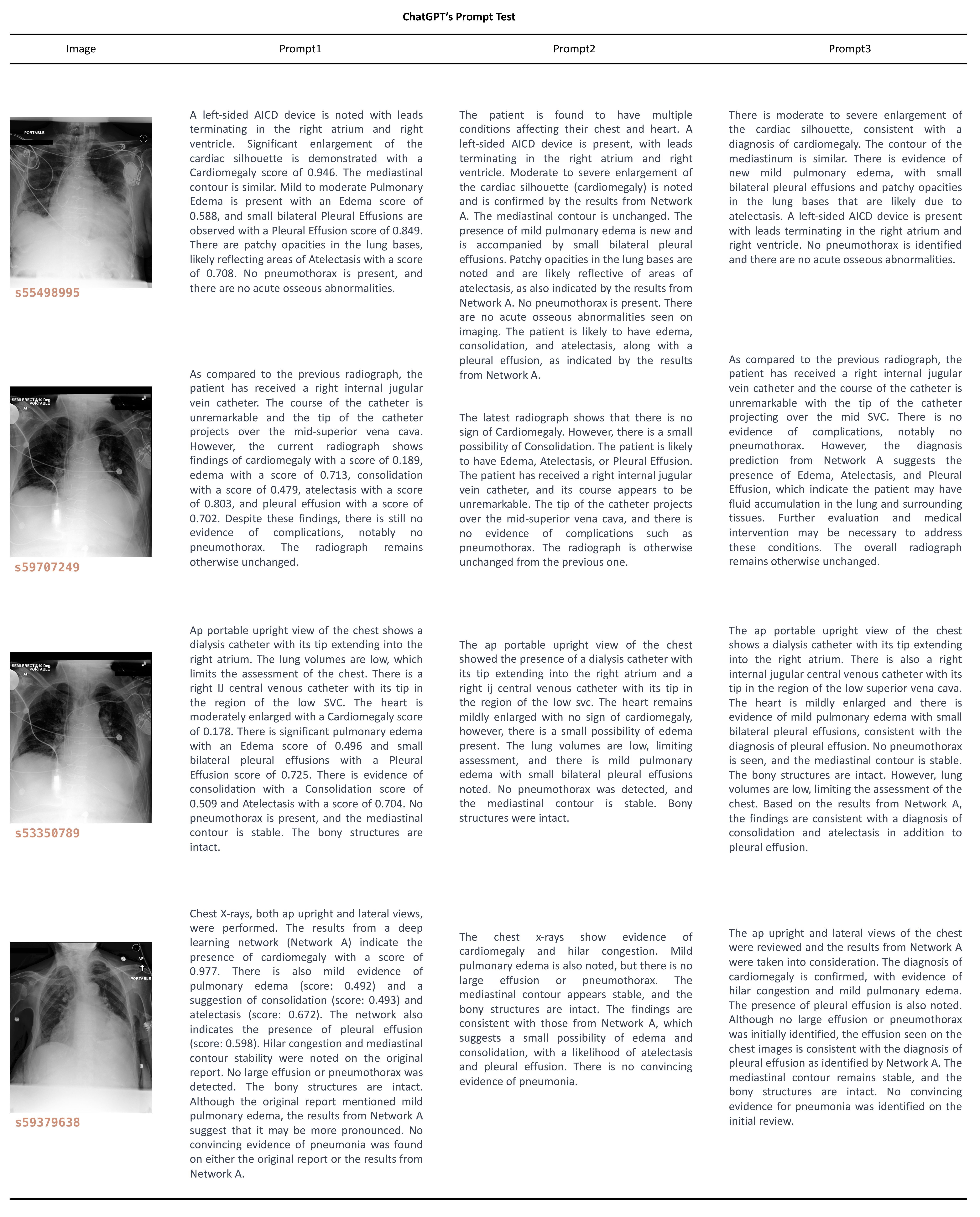}
    \caption{ChatGPT's generated reports from three different prompt designs.}
    \label{chatgpt_prompt}
\end{figure*}

\begin{figure*}[!ht]
    \centering
    \includegraphics[width=1\textwidth]{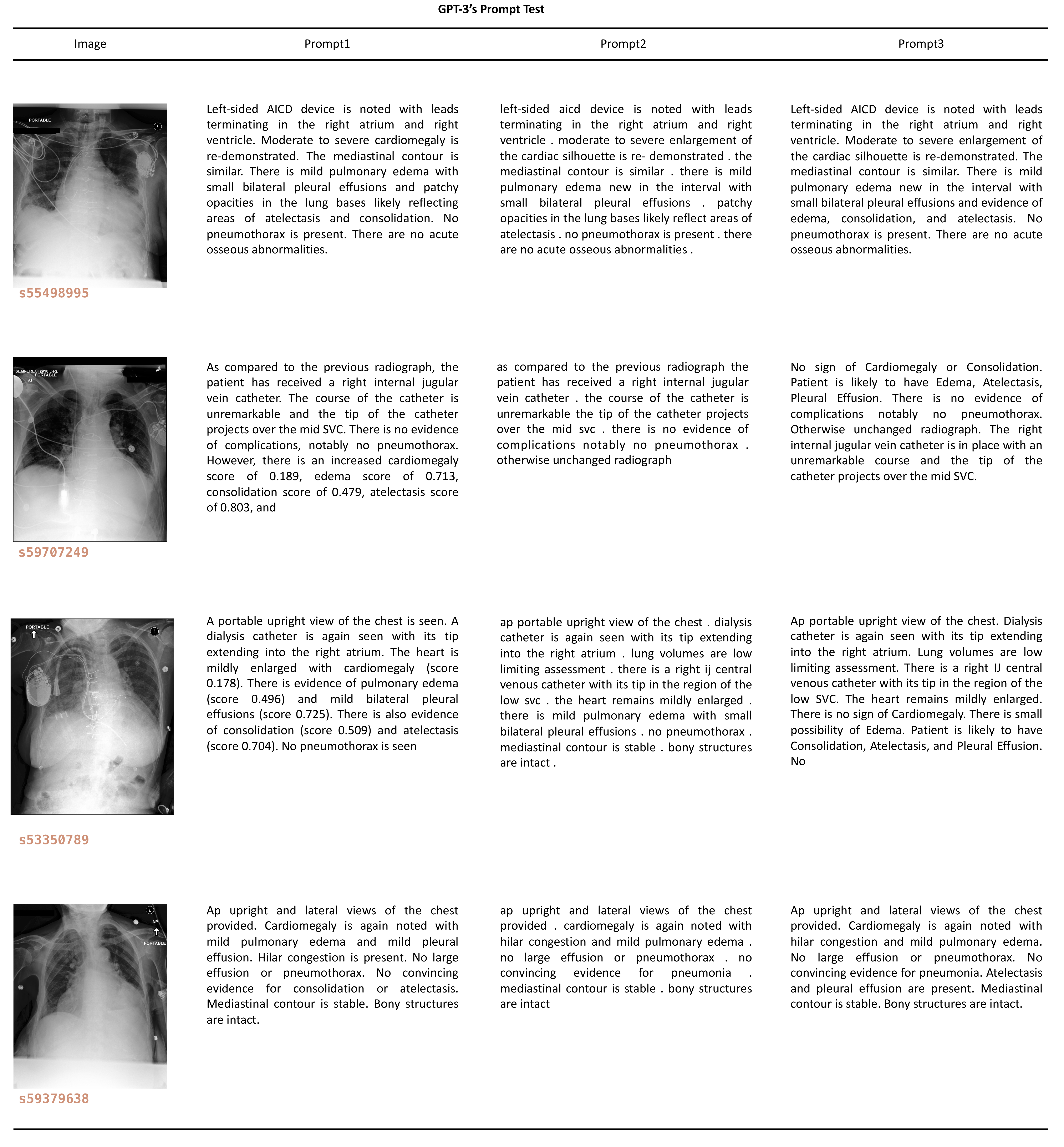}
    \caption{GPT-3's generated reports from three different prompt designs.}
    \label{gpt3_prompt}
\end{figure*}

An example is illustrated in Figure~\ref{prompt}, where the output of a disease classifier is a 5-value vector indicating the probabilities of five diseases, i.e., Cardiomegaly, Edema, Consolidation, Atelectasis, and Pleural Effusion. After that, we need to translate this result into a prompt sentence for the LLM.
A natural way of prompting is to show all five kinds of pathology and their corresponding scores. We first tell the LLM ``Higher disease score means higher possibility of illness" as the basic rule in order to avoid some misconception. Then, we represent the score of each disease as ``\$\{\textbf{disease}\} score: \$\{\textbf{score}\}" as shown in upper-right panel (Prompt\#1). Reports generated using Prompt\#1 can be found at second column in Figure~\ref{chatgpt_prompt} and Figure~\ref{gpt3_prompt}. One may notice that the LLMs are heavily influenced by Prompt\#1, usually repeating all the numbers in the output. Reports generated from Prompt\#1 are very different from radiologist's reports since the concrete diagnostic scores is not frequently used in clinical settings. 

To align with the language commonly used in clinical reports, we propose to transform the concrete scores into descriptions of disease severity as shown in lower-left panel (Prompt\#2). Prompt\#2 will be designed using a grading system, which will divide the scores into four categories: "No sign" [0.0-0.2), "Small possibility" [0.2-0.5), "Likely" [0.5-0.9), and "Definitely" [0.9 and above). These categories will be used to describe the likelihood of each of the five observations. Prompt\#3 is a concise one that reports diseases with diagnosis scores higher than 0.5 in the prompt. If no prediction is made among all five diseases, the prompt will be ``No Finding". Reports generated from Prompt\#2 and Prompt\#3 are generally acceptable and reasonable in most cases as one can observe in Figure~\ref{chatgpt_prompt} and Figure~\ref{gpt3_prompt}. 
``Network A" is frequently referenced in the generated reports. Some prompt tricks, e.g., ``\textit{Revise the report based on results from Network A but without mentioning Network A}'', can be applied to removing its mention. We do not utilize these tricks in current experiments.



\subsection{Dataset and Implementation}
In this paper, we evaluate the performance of the combination of a report generation network (R2GenCMN~\cite{chen-acl-2021-r2gencmn}) and a classification network (PCAM~\cite{ye2020weakly}). 
The report generation networks (CvT2DistilGPT2 and R2GenCMN) are trained on the MIMIC-CXR dataset~\cite{johnson2019mimic}. The MIMIC-CXR dataset is a large-scale public dataset of chest x-ray images with free-text radiology reports. It contains 377,110 images corresponding to 227,835 radiographic studies performed at the Beth Israel Deaconess Medical Center in Boston, MA. At the same time, the classifier is trained on the CheXpert dataset~\cite{irvin2019chexpert}. CheXpert is a large public dataset for chest radiograph interpretation, consisting of 224,316 chest radiographs of 65,240 patients.

The reports from the LLMs are tested on the official test set of the MIMIC dataset. Due to the current limitation of ChatGPT usage (i.e., around 20 requests per hour), we are unable to test the entire test set of MIMIC-CXR now. Therefore, 300 cases are randomly selected, including 50 cases of Cardiomegaly, 50 cases of Edema, 50 cases of Consolidation, 50 cases of Atelectasis, 50 cases of Pleural effusion, and 50 cases with no findings. During the evaluation process, the text reports were converted to multi-class labels using cheXbert~\cite{smit2020chexbert}.

The LLMs are updating constantly to include more new knowledge and events, leading to the improvement of their reasoning capability. The GPT-3 model we use in this paper is \textit{text-davinci-003} which was released by OpenAI on Feb, 2023 based on IntructGPT~\cite{ouyang2022training}. The maxlen of the output is set to 1024 and temperature set to 0.5. The ChatGPT~\cite{OpenAI2023ChatGPT} model used is the \textit{Jan-30-2023} version. 
In the section ``Interactive and Understandable CAD", ChatGPT is used to generate the example. During our test, the GPT-3 can also provide accurate and helpful chat.

\section{Report Generation}

\subsection{Quality Improvement of the Generated Report}

\begin{figure}[!ht]
    \centering
    \includegraphics[width=0.47\textwidth]{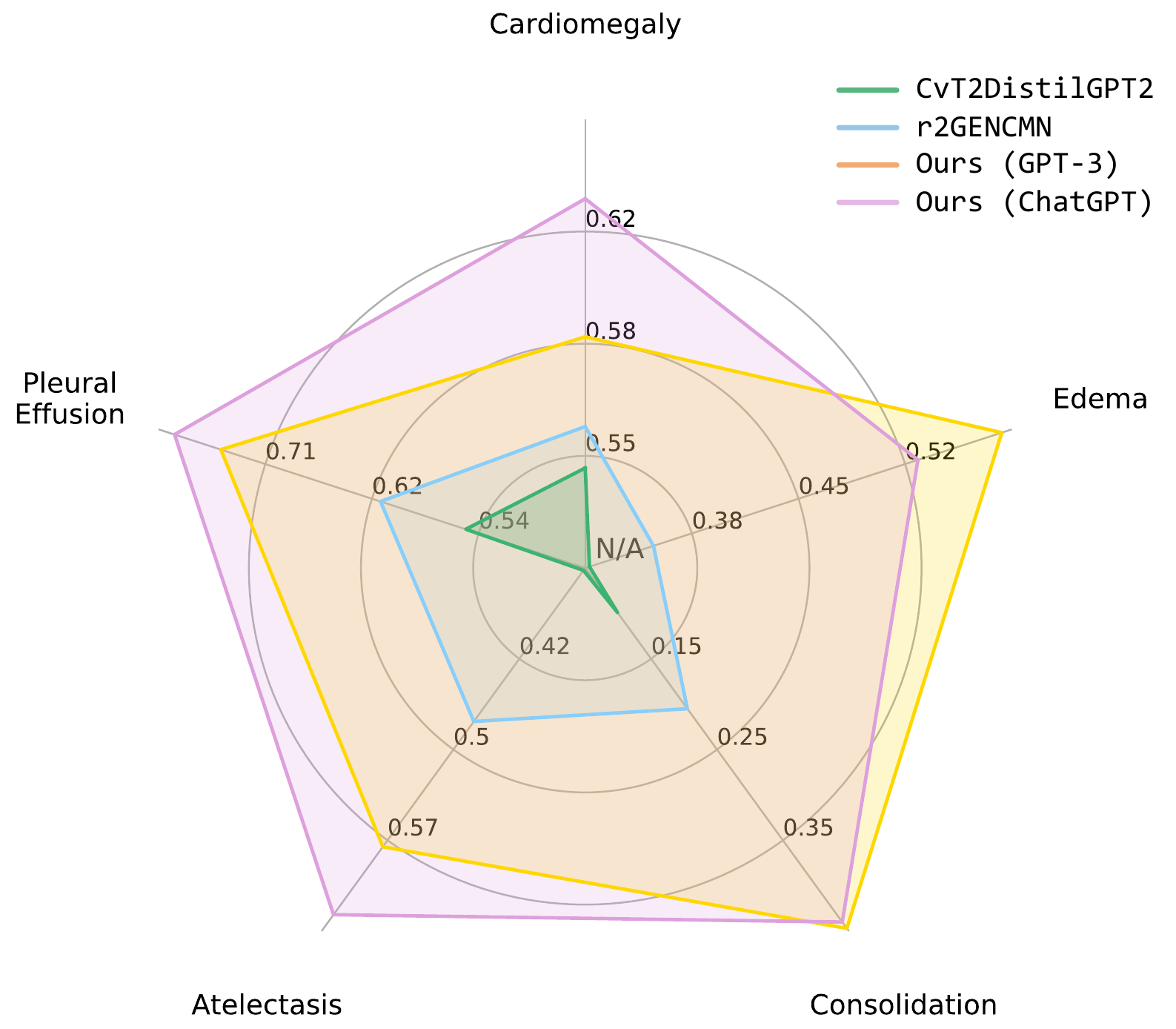}
    \caption{F1-score comparison on 5 observations.}
    \label{compare}
\end{figure}
In this section, we evaluate the performance of our proposed method with other two report-generation methods, i.e., R2GenCMN~\cite{chen-acl-2021-r2gencmn} and CvT2DistilGPT2~\cite{nicolson2022improving}. On the basis of clinical importance and prevalence, we focus on five kinds of observations. Three metrics, including precision (PR), recall (RC), and F1-score (F1), are reported in Table~\ref{tab:compare with baseline}.

The strengths of our method are clearly shown in Table~\ref{tab:compare with baseline}. It has obvious advantages in RC and F1, and is only weaker than R2GenCMN in the term of PR. Our method has a relatively high Recall and F1-score on MIMIC-CXR dataset. For all five kinds of diseases, both CvT2DistilGPT2 and R2GenCMN show inferior performance to our method concerning RC and F1. Specifically, their performances on Edema and Consolidation are rather low. Their RC values on Edema are 0.468 and 0.252, respectively, while our method achieves the RC value of 0.626 based on GPT-3. The same phenomenon can be observed in Consolidation, where the first two methods hold the values of 0.239 and 0.121 while ours (GPT-3) drastically outperforms them, with the RC value of 0.803. The R2GenCMN has a higher PR value compared to our method on three of five diseases. However, the cost of R2GenCMN's high performance on Precision is its weakness in the other two metrics, which can lead to  biased report generation, e.g., seldomly reporting any potential diseases. At the same time, our method has the highest F1 among all methods, and we believe it can be the most trustworthy report generator. 



\begin{table*}[htbp]
  \centering
  \caption{Diagnostic accuracy comparison with SOTA label generation methods. Our methods use the Prompt\#3 setting.}
    \begin{tabular}{cccccccccccccc}
    \toprule
    \multicolumn{2}{c}{\multirow{2}[4]{*}{Observations}} & \multicolumn{3}{c}{CvT2DistilGPT2~\cite{nicolson2022improving}} & \multicolumn{3}{c}{R2GenCMN~\cite{chen-acl-2021-r2gencmn}} & \multicolumn{3}{c}{Ours (GPT-3)} & \multicolumn{3}{c}{Ours (ChatGPT)} \\
\cmidrule{3-14}    \multicolumn{2}{c}{} & PR    & RC    & F1   & PR    & RC    & F1   & PR    & RC    & F1   & PR    & RC    & F1 \\
    \midrule
    \multicolumn{2}{c}{Cardiomegaly} & 0.512 & 0.591 & 0.549 & 0.590 & 0.534 & 0.561 & 0.606 & 0.569 & 0.587 & \textbf{0.663} & \textbf{0.595} & \textbf{0.627} \\
    \multicolumn{2}{c}{Edema} & 0.224 & 0.468 & 0.303 & 0.563 & 0.252 & 0.348 & \textbf{0.563} & \textbf{0.626} & 0.593 & 0.556 & 0.514 & 0.534 \\
    \multicolumn{2}{c}{Consolidation} & 0.063 & 0.239 & 0.099 & \textbf{0.667} & 0.121 & 0.205 & 0.310 & \textbf{0.803} & \textbf{0.447} & {0.322} & 0.697 & 0.440 \\
    \multicolumn{2}{c}{Atelectasis} & 0.306 & 0.388 & 0.342 & 0.442 & 0.504 & 0.471 & 0.408 & \textbf{0.991} & 0.578 & \textbf{0.470} & 0.981 & \textbf{0.636} \\
    \multicolumn{2}{c}{Pleural Effusion} & 0.454 & 0.692 & 0.548 & \textbf{0.819} & 0.500 & 0.618 & 0.634 & \textbf{0.916} & 0.749 & {0.736} & 0.845 & \textbf{0.787} \\
    \bottomrule
    \multicolumn{2}{c}{Average} & 0.312 & 0.476 & 0.368 & \textbf{0.616} & 0.382 & 0.441 & 0.504 & \textbf{0.781} & 0.591 & {0.549} & 0.726 & \textbf{0.605} \\
    \bottomrule
    \end{tabular}%
  \label{tab:compare with baseline}%
\end{table*}%

\subsection{How LLMs affect Report Quality}
In this section, we compare the performance of different LLMs for report generation. We use Prompt\#3 as the default prompt.
OpenAI provides four different sizes of GPT-3 models through its publicly accessible API: text-ada-001, text-babbage-001, text-curie-001, and text-davinci-003. The smallest text-ada-001 can not generate meaningful reports and is therefore not included in this experiment. The size of the models has not been officially disclosed. The figures listed in Table~\ref{tab:diffSizeModel} are approximate estimates based on the information in~\cite{gpt3-model-sizes}.


\begin{table*}[htbp]
  \centering
  \caption{F1-score comparison of different-size LLMs}
  \resizebox{0.8\textwidth}{!}{
    \begin{tabular}{lccccccc}
    \toprule
    Model & Size & Cardiomegaly & Edema & Consolidation & Atelectasis & Pleural Effusion & Average\\
    \midrule
    text-babbage-001 & $\sim$1.3B & 0.350 & 0.479 & 0.418 & 0.471 & 0.639 & 0.471\\
    text-curie-001 & $\sim$6.7B & 0.529 & 0.451 &  0.369 & 0.515 & 0.674 & 0.508\\
    text-davinci-003 & $\sim$175B & 0.587 & \textbf{0.593} & \textbf{0.447} & 0.578 & 0.749 & 0.591\\
    ChatGPT & $\sim$175B & \textbf{0.627} & 0.534 & 0.440 & \textbf{0.636} & \textbf{0.787} & \textbf{0.605}\\
    \bottomrule
    \end{tabular}
    }
  \label{tab:diffSizeModel}%
\end{table*}%
We report the F1-score of all observations in Table~\ref{tab:diffSizeModel}. It is noteworthy that language models struggle to perform well in clinical tasks when their model size is limited. 
The diagnostic performances of text-babbage-001 and text-curie-001 is subpar, as demonstrated by their low average F1-scores over five observations compared with the last two models. The improvement in diagnostic performance is evident in text-davinci-003, whose model size is hundreds of times larger than that of text-babbage-001. On average, text-davinci-003's F1-score is improved from 0.471 to 0.591. The ChatGPT is slightly better than text-davinci-003, achieving the improvement of 0.014, and their diagnostic abilities are comparable. Overall, the diagnostic capability of language models is proportional to their size, highlighting the critical role of the logistic reasoning capability of LLMs.

\begin{figure}[!ht]
    \centering
    \includegraphics[width=0.45\textwidth]{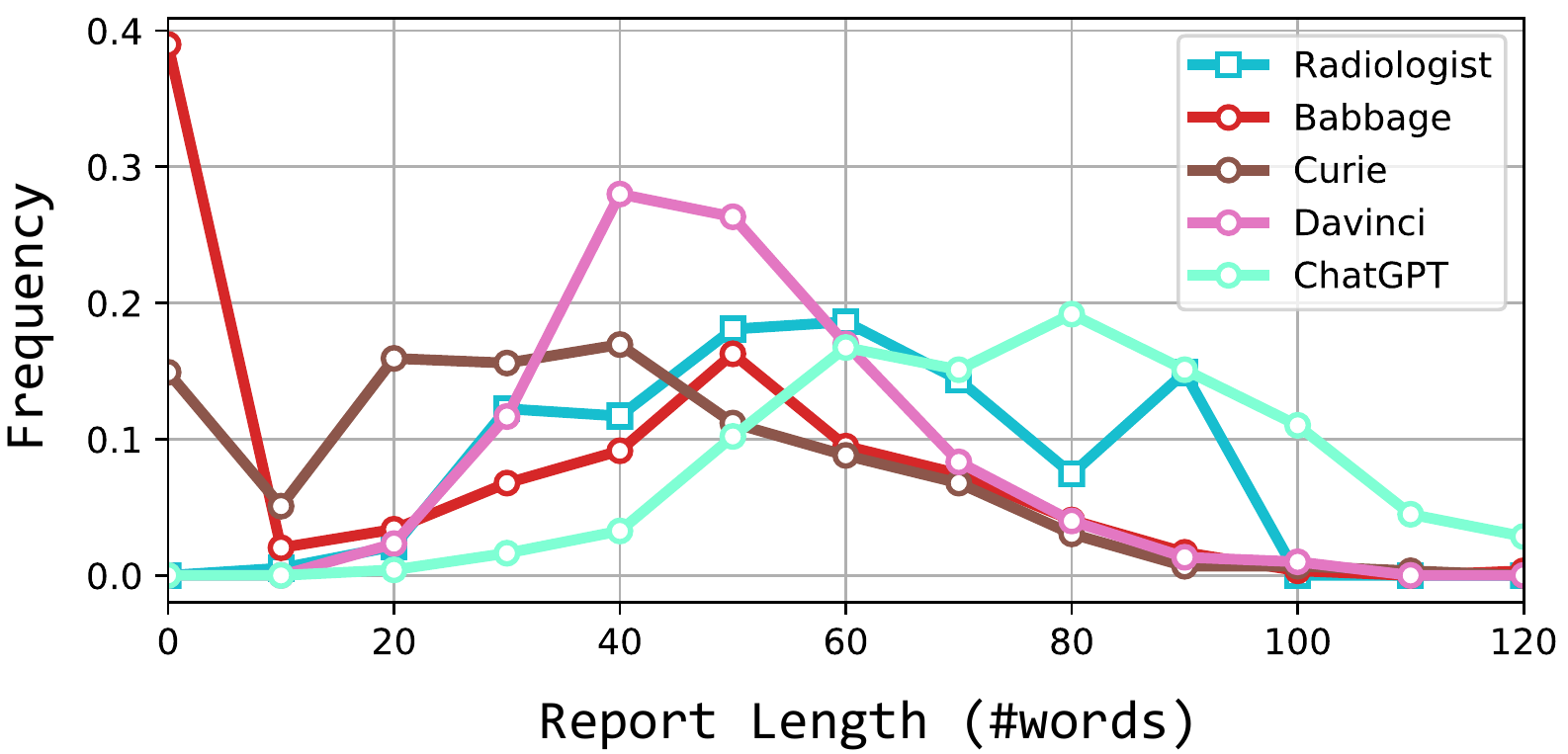}
    \caption{Length of Reports generated by different models. ``Babbage", ``Curie" and ``Davinci" represent the three GPT-3 models with different model sizes, i.e., text-babbage-001, text-curie-001, and text-davinci-003.}
    \label{len_model}
\end{figure}

In our experiments, it can be observed that more capable models generally produce longer reports as shown in Figure~\ref{len_model}.
At the same time, nearly 40\% of reports generated by text-babbage-001 and nearly 15\% of reports generated by text-curie-001 have no meaningful content. 

\section{Interactive and Understandable CAD}

\begin{figure*}[!ht]
    \centering
    \includegraphics[width=1\textwidth]{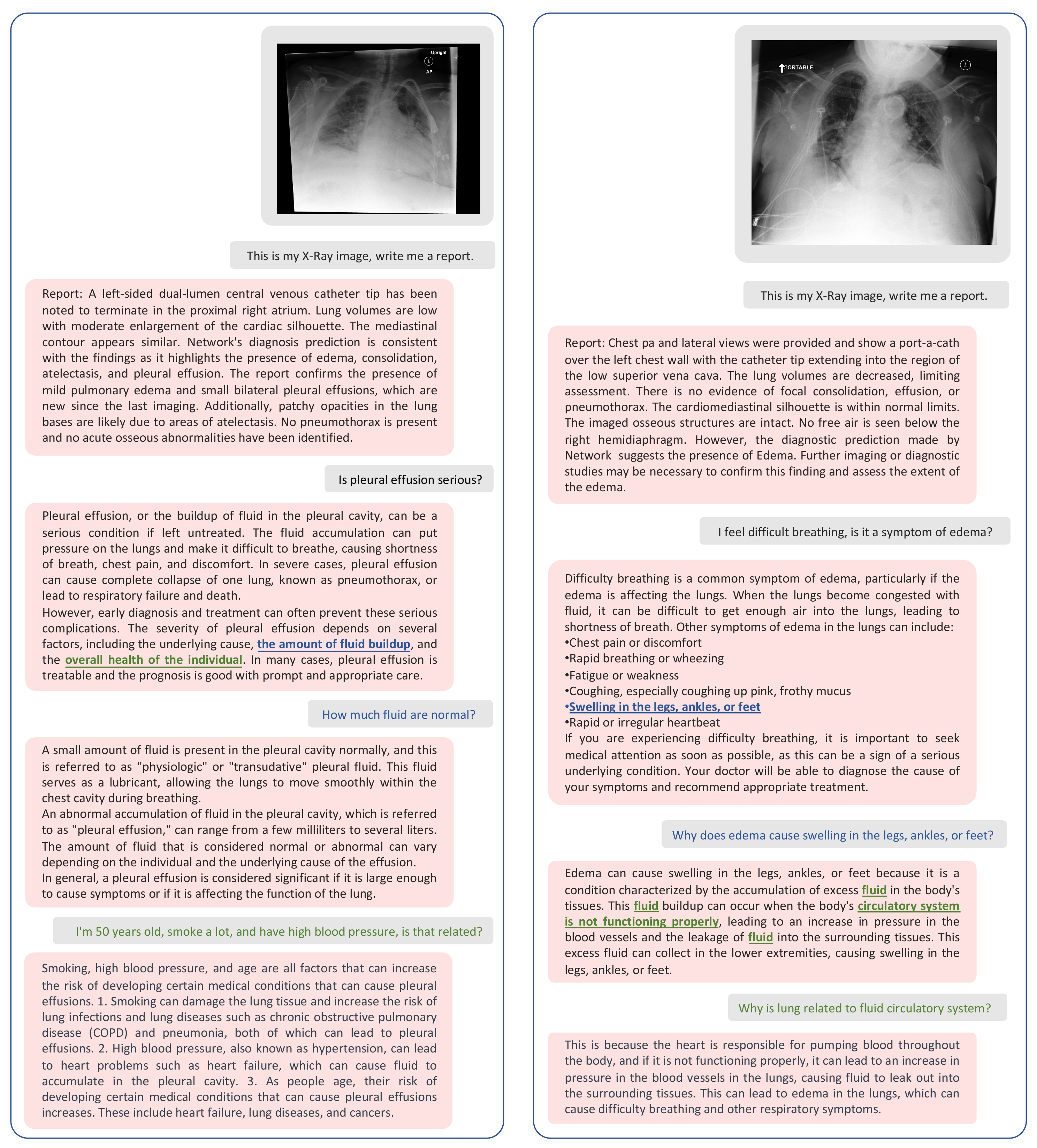}
    \caption{Two examples of interactive CAD using our framework and ChatGPT.}
    \label{interactive}
\end{figure*}

The proposed ChatCAD offers several benefits, including its ability to utilize LLM's extensive and reliable medical knowledge to provide interactive explanations and advice. As shown in Figure~\ref{interactive}, two examples of the interactive CAD are provided, with one chat discussing pleural effusion and the other addressing edema and its relationship to swelling.

Through this approach, patients can gain a clearer understanding of their symptoms, diagnosis, and treatment options, leading to more efficient and cost-effective consultations with medical experts. As language models continue to advance and become more accurate with access to more trustworthy medical training data, ChatCAD has the potential to significantly enhance the quality of online healthcare services.

\section{Limitations and Discussion}
In this paper, we explore a novel framework, ChatCAD, introducing large language models in CAD. The proposed method, however, still has limitations to be solved.

First, LLM-generated reports are not human like in a certain way. LLM is likely to output sentences like ``Network A's diagnosis prediction is consistent with the findings in the radiological report" or ``The findings from Network A's diagnosis prediction are supported by the X-ray". This is reflected on natural language similarity metrics when we compare to our baseline method. ChatCAD improved the diagnosis accuracy but dropped the BLEU score~\cite{papineni2002bleu}. A promising way to address this issue is to add a module after ChatGPT to filter generated reports. Or add prompt like ``please do not mention Network A".

Additionally, we only design three typical kinds of prompts that are intuitive, and there is room for improvement. LLMs are capable of solving logical reasoning problems without additional computational costs ~\cite{wei2022chain}. 
In current ChatCAD, we did not give the network about patient's major complaint since there is no such dataset available. 
We believe the LLMs can process more complex
information than what we currently provide. Better datasets and benchmarks are needed.

Our experiments demonstrate significant impact of language model size on diagnostic accuracy. Larger, advanced, and more truthy LLMs such as the upcoming GPT-4 may improve the accuracy and report quality further. However, the role of vision classifiers has not yet been explored, and additional research is necessary to determine if models such as ViT~\cite{dosovitskiy2020image} or SwinTransformer~\cite{liu2021swin}, which boast larger parameters, can deliver improved results. On the other hand, LLMs can also be used to help the training of vision models, such as correcting outputs of vision models using related medical knowledge learned in LLMs.

In our work, we have only carried out a qualitative analysis of the prompt design instead of a quantitative analysis. Further in-depth investigations will be undertaken once the API for ChatGPT becomes available for use. Moreover, the specifics of this paper have not been discussed with any clinical professionals, and therefore it still lacks rigor in many places. We will improve it in subsequent versions.




{\small
\bibliographystyle{ieee_fullname}
\bibliography{PaperForReview}

\begin{thebibliography}{10}\itemsep=-1pt

\bibitem{alayrac2022flamingo}
Jean-Baptiste Alayrac, Jeff Donahue, Pauline Luc, Antoine Miech, Iain Barr,
  Yana Hasson, Karel Lenc, Arthur Mensch, Katie Millican, Malcolm Reynolds,
  et~al.
\newblock Flamingo: a visual language model for few-shot learning.
\newblock {\em arXiv preprint arXiv:2204.14198}, 2022.

\bibitem{alsentzer2019publicly}
Emily Alsentzer, John~R Murphy, Willie Boag, Wei-Hung Weng, Di Jin, Tristan
  Naumann, and Matthew McDermott.
\newblock Publicly available clinical bert embeddings.
\newblock {\em arXiv preprint arXiv:1904.03323}, 2019.

\bibitem{brown2020language}
Tom Brown, Benjamin Mann, Nick Ryder, Melanie Subbiah, Jared~D Kaplan, Prafulla
  Dhariwal, Arvind Neelakantan, Pranav Shyam, Girish Sastry, Amanda Askell,
  et~al.
\newblock Language models are few-shot learners.
\newblock {\em Advances in neural information processing systems},
  33:1877--1901, 2020.

\bibitem{chen-acl-2021-r2gencmn}
Zhihong Chen, Yaling Shen, Yan Song, and Xiang Wan.
\newblock Generating radiology reports via memory-driven transformer.
\newblock In {\em Proceedings of the Joint Conference of the 59th Annual
  Meeting of the Association for Computational Linguistics and the 11th
  International Joint Conference on Natural Language Processing}, Aug. 2021.

\bibitem{devlin2018bert}
Jacob Devlin, Ming-Wei Chang, Kenton Lee, and Kristina Toutanova.
\newblock Bert: Pre-training of deep bidirectional transformers for language
  understanding.
\newblock {\em arXiv preprint arXiv:1810.04805}, 2018.

\bibitem{dosovitskiy2020image}
Alexey Dosovitskiy, Lucas Beyer, Alexander Kolesnikov, Dirk Weissenborn,
  Xiaohua Zhai, Thomas Unterthiner, Mostafa Dehghani, Matthias Minderer, Georg
  Heigold, Sylvain Gelly, et~al.
\newblock An image is worth 16x16 words: Transformers for image recognition at
  scale.
\newblock {\em arXiv preprint arXiv:2010.11929}, 2020.

\bibitem{gpt3-model-sizes}
Leo Gao.
\newblock On the sizes of openai api models.

\bibitem{gu2021domain}
Yu Gu, Robert Tinn, Hao Cheng, Michael Lucas, Naoto Usuyama, Xiaodong Liu,
  Tristan Naumann, Jianfeng Gao, and Hoifung Poon.
\newblock Domain-specific language model pretraining for biomedical natural
  language processing.
\newblock {\em ACM Transactions on Computing for Healthcare (HEALTH)},
  3(1):1--23, 2021.

\bibitem{herdade2019image}
Simao Herdade, Armin Kappeler, Kofi Boakye, and Joao Soares.
\newblock Image captioning: Transforming objects into words.
\newblock {\em Advances in neural information processing systems}, 32, 2019.

\bibitem{irvin2019chexpert}
Jeremy Irvin, Pranav Rajpurkar, Michael Ko, Yifan Yu, Silviana Ciurea-Ilcus,
  Chris Chute, Henrik Marklund, Behzad Haghgoo, Robyn Ball, Katie Shpanskaya,
  et~al.
\newblock Chexpert: A large chest radiograph dataset with uncertainty labels
  and expert comparison.
\newblock In {\em Proceedings of the AAAI conference on artificial
  intelligence}, volume~33, pages 590--597, 2019.

\bibitem{johnson2019mimic}
Alistair~EW Johnson, Tom~J Pollard, Seth~J Berkowitz, Nathaniel~R Greenbaum,
  Matthew~P Lungren, Chih-ying Deng, Roger~G Mark, and Steven Horng.
\newblock Mimic-cxr, a de-identified publicly available database of chest
  radiographs with free-text reports.
\newblock {\em Scientific data}, 6(1):317, 2019.

\bibitem{kung2022performance}
Tiffany~H Kung, Morgan Cheatham, Arielle Medinilla, ChatGPT, Czarina Sillos,
  Lorie De~Leon, Camille Elepano, Marie Madriaga, Rimel Aggabao, Giezel
  Diaz-Candido, et~al.
\newblock Performance of chatgpt on usmle: Potential for ai-assisted medical
  education using large language models.
\newblock {\em medRxiv}, pages 2022--12, 2022.

\bibitem{lee2020biobert}
Jinhyuk Lee, Wonjin Yoon, Sungdong Kim, Donghyeon Kim, Sunkyu Kim, Chan~Ho So,
  and Jaewoo Kang.
\newblock Biobert: a pre-trained biomedical language representation model for
  biomedical text mining.
\newblock {\em Bioinformatics}, 36(4):1234--1240, 2020.

\bibitem{liu2019clinically}
Guanxiong Liu, Tzu-Ming~Harry Hsu, Matthew McDermott, Willie Boag, Wei-Hung
  Weng, Peter Szolovits, and Marzyeh Ghassemi.
\newblock Clinically accurate chest x-ray report generation.
\newblock In {\em Machine Learning for Healthcare Conference}, pages 249--269.
  PMLR, 2019.

\bibitem{liu2021swin}
Ze Liu, Yutong Lin, Yue Cao, Han Hu, Yixuan Wei, Zheng Zhang, Stephen Lin, and
  Baining Guo.
\newblock Swin transformer: Hierarchical vision transformer using shifted
  windows.
\newblock In {\em Proceedings of the IEEE/CVF International Conference on
  Computer Vision}, pages 10012--10022, 2021.

\bibitem{nicolson2022improving}
Aaron Nicolson, Jason Dowling, and Bevan Koopman.
\newblock Improving chest x-ray report generation by leveraging warm-starting.
\newblock {\em arXiv preprint arXiv:2201.09405}, 2022.

\bibitem{OpenAI2023ChatGPT}
OpenAI.
\newblock Chatgpt: Optimizing language models for dialogue, 2023.

\bibitem{ouyang2022training}
Long Ouyang, Jeff Wu, Xu Jiang, Diogo Almeida, Carroll~L Wainwright, Pamela
  Mishkin, Chong Zhang, Sandhini Agarwal, Katarina Slama, Alex Ray, et~al.
\newblock Training language models to follow instructions with human feedback.
\newblock {\em arXiv preprint arXiv:2203.02155}, 2022.

\bibitem{ouyang2020dual}
Xi Ouyang, Jiayu Huo, Liming Xia, Fei Shan, Jun Liu, Zhanhao Mo, Fuhua Yan,
  Zhongxiang Ding, Qi Yang, Bin Song, et~al.
\newblock Dual-sampling attention network for diagnosis of covid-19 from
  community acquired pneumonia.
\newblock {\em IEEE Transactions on Medical Imaging}, 39(8):2595--2605, 2020.

\bibitem{papineni2002bleu}
Kishore Papineni, Salim Roukos, Todd Ward, and Wei-Jing Zhu.
\newblock Bleu: a method for automatic evaluation of machine translation.
\newblock In {\em Proceedings of the 40th annual meeting of the Association for
  Computational Linguistics}, pages 311--318, 2002.

\bibitem{radford2021learning}
Alec Radford, Jong~Wook Kim, Chris Hallacy, Aditya Ramesh, Gabriel Goh,
  Sandhini Agarwal, Girish Sastry, Amanda Askell, Pamela Mishkin, Jack Clark,
  et~al.
\newblock Learning transferable visual models from natural language
  supervision.
\newblock In {\em International Conference on Machine Learning}, pages
  8748--8763. PMLR, 2021.

\bibitem{radford2018improving}
Alec Radford, Karthik Narasimhan, Tim Salimans, Ilya Sutskever, et~al.
\newblock Improving language understanding by generative pre-training.
\newblock 2018.

\bibitem{raffel2020exploring}
Colin Raffel, Noam Shazeer, Adam Roberts, Katherine Lee, Sharan Narang, Michael
  Matena, Yanqi Zhou, Wei Li, and Peter~J Liu.
\newblock Exploring the limits of transfer learning with a unified text-to-text
  transformer.
\newblock {\em The Journal of Machine Learning Research}, 21(1):5485--5551,
  2020.

\bibitem{shen2017deep}
Dinggang Shen, Guorong Wu, and Heung-Il Suk.
\newblock Deep learning in medical image analysis.
\newblock {\em Annual review of biomedical engineering}, 19:221--248, 2017.

\bibitem{singhal2022large}
Karan Singhal, Shekoofeh Azizi, Tao Tu, S~Sara Mahdavi, Jason Wei, Hyung~Won
  Chung, Nathan Scales, Ajay Tanwani, Heather Cole-Lewis, Stephen Pfohl, et~al.
\newblock Large language models encode clinical knowledge.
\newblock {\em arXiv preprint arXiv:2212.13138}, 2022.

\bibitem{smit2020chexbert}
Akshay Smit, Saahil Jain, Pranav Rajpurkar, Anuj Pareek, Andrew~Y Ng, and
  Matthew~P Lungren.
\newblock Chexbert: combining automatic labelers and expert annotations for
  accurate radiology report labeling using bert.
\newblock {\em arXiv preprint arXiv:2004.09167}, 2020.

\bibitem{tsimpoukelli2021multimodal}
Maria Tsimpoukelli, Jacob~L Menick, Serkan Cabi, SM Eslami, Oriol Vinyals, and
  Felix Hill.
\newblock Multimodal few-shot learning with frozen language models.
\newblock {\em Advances in Neural Information Processing Systems}, 34:200--212,
  2021.

\bibitem{vaswani2017attention}
Ashish Vaswani, Noam Shazeer, Niki Parmar, Jakob Uszkoreit, Llion Jones,
  Aidan~N Gomez, {\L}ukasz Kaiser, and Illia Polosukhin.
\newblock Attention is all you need.
\newblock In {\em Advances in neural information processing systems}, pages
  5998--6008, 2017.

\bibitem{wang2022crossPro}
Jun Wang, Abhir Bhalerao, and Yulan He.
\newblock Cross-modal prototype driven network for radiology report generation.
\newblock In {\em Computer Vision--ECCV 2022: 17th European Conference, Tel
  Aviv, Israel, October 23--27, 2022, Proceedings, Part XXXV}, pages 563--579.
  Springer, 2022.

\bibitem{wang2022follow}
Sheng Wang, Xi Ouyang, Tianming Liu, Qian Wang, and Dinggang Shen.
\newblock Follow my eye: Using gaze to supervise computer-aided diagnosis.
\newblock {\em IEEE Transactions on Medical Imaging}, 2022.

\bibitem{wei2022chain}
Jason Wei, Xuezhi Wang, Dale Schuurmans, Maarten Bosma, Ed Chi, Quoc Le, and
  Denny Zhou.
\newblock Chain of thought prompting elicits reasoning in large language
  models.
\newblock {\em arXiv preprint arXiv:2201.11903}, 2022.

\bibitem{ye2020weakly}
Wenwu Ye, Jin Yao, Hui Xue, and Yi Li.
\newblock Weakly supervised lesion localization with probabilistic-cam pooling,
  2020.

\bibitem{you2016image}
Quanzeng You, Hailin Jin, Zhaowen Wang, Chen Fang, and Jiebo Luo.
\newblock Image captioning with semantic attention.
\newblock In {\em Proceedings of the IEEE conference on computer vision and
  pattern recognition}, pages 4651--4659, 2016.

\bibitem{zhang2020radiologyConGraph}
Yixiao Zhang, Xiaosong Wang, Ziyue Xu, Qihang Yu, Alan Yuille, and Daguang Xu.
\newblock When radiology report generation meets knowledge graph.
\newblock In {\em Proceedings of the AAAI Conference on Artificial
  Intelligence}, volume~34, pages 12910--12917, 2020.

\bibitem{zhao2023rcps}
Xiangyu Zhao, Zengxin Qi, Sheng Wang, Qian Wang, Xuehai Wu, Ying Mao, and Lichi
  Zhang.
\newblock Rcps: Rectified contrastive pseudo supervision for semi-supervised
  medical image segmentation.
\newblock {\em arXiv preprint arXiv:2301.05500}, 2023.

\end{thebibliography}
}
\newpage

\end{document}